\documentclass[lettersize,journal]{IEEEtran}
\usepackage{amsmath,amsfonts}
\usepackage{algorithmic}
\usepackage{algorithm}
\usepackage{array}
\usepackage[caption=false,font=normalsize,labelfont=sf,textfont=sf]{subfig}
\usepackage{textcomp}
\usepackage{stfloats}
\usepackage{url}
\usepackage{verbatim}
\usepackage{graphicx}
\usepackage{cite}
\usepackage{caption}
\usepackage{url}
\usepackage{hyperref} 
\usepackage{multirow}
\usepackage{makecell}
\usepackage{listings}
\usepackage[flushleft]{threeparttable}
\usepackage{booktabs}
\usepackage{siunitx}
\usepackage{subfig}
\usepackage{xcolor}
\usepackage[noabbrev, capitalise]{cleveref}

\begin{document}

\title{EI-Drive: A Platform for Cooperative Perception with Realistic Communication Models}

\author{Hanchu Zhou, Edward Xie, Wei Shao, Dechen Gao, Michelle Dong, Junshan Zhang \IEEEmembership{Fellow, IEEE}

\thanks{Manuscript received April xx, xxxx; revised August xx, xxxx. \textit{(Corresponding author: Junshan Zhang.)}}

\thanks{Hanchu Zhou, Dechen Gao, and Junshan Zhang are with the Davis AI, Robotics, and Edge Laboratory, University of California, Davis, CA 95616 USA (e-mail: hczhou@ucdavis.edu; dcgao@ucdavis.edu; jazh@ucdavis.edu). }
\thanks{Edward Xie is with the Electrical and Computer Engineering Department of Johns Hopkins University, Baltimore, MD 21218 USA. (e-mail: exie3@jh.edu)}
\thanks{Wei Shao is with data61, CSIRO, Clayton, Victoria, Australia. (e-mail: phdweishao@gmail.com)}
\thanks{Michelle Dong is with Monta Vista High School, Cupertino, CA 95014 USA. (e-mail: dongmichelley@gmail.com)}
}

\markboth{Journal of \LaTeX\ Class Files,~Vol.~14, No.~8, August~2024}%
{Shell \MakeLowercase{\textit{et al.}}: A Sample Article Using IEEEtran.cls for IEEE Journals}

\IEEEpubid{0000--0000/00\$00.00~\copyright~2024 IEEE}

\maketitle

\begin{abstract}
The growing interest in autonomous driving calls for realistic simulation platforms capable of accurately simulating cooperative perception process in realistic traffic scenarios. Existing studies for cooperative perception often have not accounted for transmission latency and errors in real-world environments. To address this gap, we introduce EI-Drive, an edge-AI based autonomous driving simulation platform that integrates advanced cooperative perception with   more realistic communication models. Built on the CARLA framework, EI-Drive features new modules for cooperative perception while  taking into account transmission latency and errors, providing a more realistic platform for evaluating cooperative perception algorithms. In particular,  the platform enables vehicles to fuse data from multiple sources, improving situational awareness and safety in complex environments. With its modular design, EI-Drive allows for detailed exploration of sensing, perception, planning, and control in various cooperative driving scenarios. Experiments using EI-Drive demonstrate significant improvements in vehicle safety and performance, particularly in scenarios with complex traffic flow and network conditions. All code and documents are accessible on our GitHub page: \url{https://ucd-dare.github.io/eidrive.github.io/}.
\end{abstract}


\section{Introduction}
Simulation platforms play a crucial role in the development and testing of autonomous vehicles (AVs) since AV systems must navigate various complex environments to validate their safety and reliability. However, conducting such evaluation in the real world can be dangerous, inefficient, and prohibitively expensive. To address these challenges, simulation platforms serve as alternative mechanisms to provide controllable, realistic, and cost-effective environments where AV algorithms can be evaluated across various scenarios, from simple maneuvers to complex traffic interactions~\cite{rosique2019systematic}.

Cooperative perception, with growing attention in autonomous driving, enables multiple vehicles or Road Side Units (RSUs) to share sensor data with each other and enhance their collective understanding of the environment~\cite{kara2020effect}. This approach is critical for overcoming limitations like sensor occlusions, limited fields of view, and noise, which can lead to missed obstacles or incorrect decisions~\cite{xu2022cobevt}. 

Existing platforms have developed functionalities on cooperative perception and enabled their interactions with simulation environments.
A major limitation of current AV simulation platforms has root in their use of unrealistic communication models. Since cooperative perception hinges heavily on communications between agents, data transmission  plays a crucial role in the quality of perception results received. Critical factors like transmission latency and errors can negatively impact real-time decision-making, thereby impairing the overall performance. Most of the existing studies have not accounted for transmission latency and errors in cooperative perception, both of which however are inevitable in real-world vehicular networks~\cite{ashraf2017towards, shao2024impact}.

\IEEEpubidadjcol

Aiming to  evaluate the performance and robustness of autonomous driving algorithms in real-world communication conditions, we develop EI-Drive to incorporate realistic communication models into simulation platform design.
The proposed EI-Drive platform integrates innovative features specifically designed to overcome the limitation of existing AV simulation platforms, as outlined below. First, EI-Drive incorporates a realistic communication model that simulates latency and errors during data transmission.  By accounting for transmission latency and errors, the platform ensures that AV systems can be evaluated under conditions that mirror real-world networks conditions. Second, EI-Drive introduces comprehensive support for advanced cooperative perception by enabling data fusion from heterogeneous agents, such as vehicles and RSUs, in various tasks. This allows vehicles to share and fuse sensory information, overcoming occlusions and sensor limitations to improve environmental awareness. Lastly, EI-Drive is designed to simulate complex multi-agent environments, enabling thorough testing of AV performance in dynamic, high-risk situations. This includes customizable driving scenarios with interactions between multiple vehicles, pedestrians, and RSUs, ensuring robust evaluations of autonomous algorithms. These innovations make EI-Drive a  promising  platform for advancing AV research and development.

\begin{figure*}
    \centering
    \includegraphics[width=\linewidth]{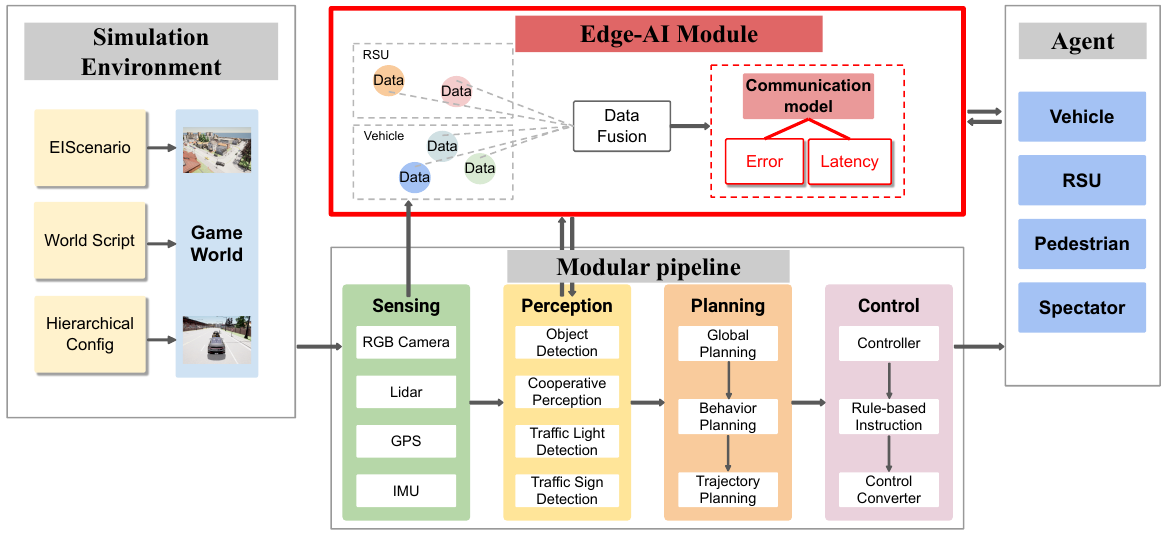}
    \caption{The framework of EI-Drive, which consists of four main components: simulation environment, edge-AI module, modular pipeline, and agent.}
    \label{fig:method_architecture}
\end{figure*}

The main contributions of this study are summarized as follows:  

\begin{enumerate}
    \item \textbf{Design of EI-Drive}: The paper introduces EI-Drive, an edge-AI based simulation platform that integrates realistic communication models into the design of cooperative perception. This allows multiple agents, such as vehicles and RSUs, to share and fuse sensor data, significantly enhancing their situational awareness in complex driving environments. In particular, communication models that account for transmission latency and errors provides an accurate representation of the real-world network conditions faced by AVs, improving the robustness and reliability in AV testing.

    \item \textbf{Modules for modeling the interactions between heterogeneous agents via customizable built-in scenarios}: EI-Drive enhances simulation capabilities for highly dynamic, multi-agent environments, offering customizable scenarios for evaluation in complex traffic and varying network conditions. These built-in scenarios, integrating world scripts and EIScenarios, are tailored to explore the interactions between heterogeneous agents and various functionalities, such as cooperative perception and communication models.  facilitating the testing of autonomous algorithms in diverse challenging environments.

    \item \textbf{Extensive experimental studies of cooperative perception under transmission latency and errors}: Based on the built-in scenarios, we conduct experiments of cooperative perception with realistic communication models. The results underscore the significant impact of transmission latency and errors on the performance of cooperative perception and the vehicle behavior. Furthermore, the experiments highlight EI-Drive's capabilities as a powerful tool, effectively simulating the intricate interactions between communication models and autonomous driving system.
\end{enumerate}

\section{Related Work}
\label{sec:related}
\subsection{Autonomous driving simulation platform}
Testing autonomous vehicles in the real world is costly and risky, making virtual simulation essential for evaluating algorithms before deployment. Therefore, autonomous driving simulation platforms are developed to provide a controlled environment for prototyping and testing. CARLA \cite{dosovitskiy2017carla} is an open-source simulator famous for its highly realistic urban environment and built-in functions for traffic management. Focusing on traffic management, SUMO \cite{lopez2018microscopic} is designed for simulating microscopic and large-scale traffic within road network. Based on SUMO, FLOW \cite{wu2021flow} aims to optimize the traffic control using Reinforcement Learning (RL). Furthermore, some simulation platforms are  designed to support specific research areas in autonomous driving. To provide an easy-to-use World-Model-based autonomous driving platform, CarDreamer \cite{gao2024cardreamer} integrates World Model with built-in scenarios for fast training. Autoware \cite{autoware} provides full-stack simulation from sensing to control based on Robot Operating System (ROS). To better test RL algorithm on integrated environments, MetaDrive \cite{li2022metadrive} constructs a variety of RL tasks and baselines in both single-agent and multi-agent settings. AVstack \cite{hallyburton2023avstack} aims to maintain the compatibility with multiple open-source AV algorithms across metrics, addressing challenges in testing across different datasets and simulators. To our best knowledge, there is no open-source platform currently provide realistic communication features to account for transmission errors and latency in cooperative perception tasks. By integrating communication model with cooperative perception, EI-Drive aims to filled this gap  for cooperative driving automation.

\subsection{Cooperative perception}
To reduce the weaknesses of single-vehicle perception, such as obstruction, limited view, and sensor error \cite{huang2023v2x}, cooperative perception fuses the data from multiple agents and generates a more accurate representation of the environment. Based on the data fusion methods, cooperative perception can be  classified into three categories: 1) early fusion \cite{jia2022online} \cite{jia2023mass}, where the raw data is combined before  processing; 2) intermediate fusion \cite{wang2022collaborative} \cite{chi2023multimodal}, where the fusion occurs after a certain level of feature extraction has been performed; 3) late fusion \cite{aoki2020cooperative} \cite{zhou2022aicp}, where the result of each agent's perception module is fused. OpenCDA \cite{xu2021opencda}, as a platform for cooperative perception, has been developed based on CARLA to provide a framework for cooperative perception in platooning. While OpenCDA involves information sharing among agents, it does not include the important transmission features, such as latency and error, which would have significant impact on the performance of cooperative perception.

\IEEEpubidadjcol

\subsection{Communication}
Several simulation platforms have been designed for the research in network communication. ns-3 \cite{riley2010ns} is one of the most widely used discrete-event network simulators. Its modular design and wide protocol support allow easy extension in various experiments. OMNeT++ \cite{varga2010overview} is a versatile and extensible simulation framework for communication networks. Based on its great extensibility, several simulators are developed on OMNeT++ framework, including Veins \cite{sommer2019veins} and Simu5G \cite{nardini2020simu5g}. Since ns-3 and OMNeT++ are developed based on C++, it is difficult to build an interface to integrate with the mainstream Python-based autonomous driving simulators, which makes AV research with communication challenging. Though there are some communication simulators feature V2X communication, such as VSimRTI \cite{schunemann2011v2x} and platform in \cite{cuiintegration}, they are either implemented on an unrealistic 2D environment or do not involve cooperative perception.

\section{The EI-Drive Framework}

This section presents a comprehensive overview of EI-Drive and its main components, and provides details for  the important features of  different modules as well as  the relations between these modules.

\subsection{Simulation environment}\label{sub:environment}
EI-Drive is designed to provide customizable simulation environments that are convenient to manage. The simulation environment has three key components: GameWorld, world script, and EIScenario.

\textbf{GameWorld.} The simulation environment is established on GameWorld, which is developed base on CARLA \cite{CARLAdocumentation} environment and integrates agents, maps, and   simulation configurations. The GameWorld offers a variety of methods to spawn agents, including by specific location, by range, and by list. This flexibility allows users to tailor the spawning of a single vehicle or heavy background traffic to meet experiment demands. 

Additionally, GameWorld provides an API to set the weather in simulation, allowing users to adjust parameters such as the sun altitude angle, cloudiness, and fog density. The pre-configured weather can be simply loaded by a YAML file, enabling experiments under different weather conditions.

\textbf{World script.} The world script manages behaviors and events in a designed scenario. The events with trigger conditions that influence the agents' behaviors can be defined in the script. A specific example is the implementation of cooperative perception, where the participants, tasks, and methods of cooperative perception are specified in the world script. Comparing to GameWorld, which handles the fundamental aspects of environment and agents, the world script manages the details of events, allowing for customization in simulation.

Besides, world script is responsible for visualization. Our SpectatorController provides various pre-built spectator movements, including follow and still modes, which can be easily configured to monitor the simulation comprehensively. In some complex tasks, such as cooperative perception, Pygame is utilized to enhance the visualization of perception results, where the information from CARLA is projected onto Pygame interface to present the results clearly.

\textbf{EIScenario.} EIScenario is designed to make EI-Drive compatible with ScenarioRunner, a module to facilitate the creation, execution and evaluation of driving scenarios. It offers predefined scenarios and enables users to define customized scenarios with Python or OpenSCENARIO \cite{ASAMopenscenario}, an open standard for describing driving scenarios. This provides user with an additional approach to defining and managing scenarios.

EIScenario script is built on top of ScenarioRunner. It has a similar structure to world script, while the environment and agent behavior are handled by ScenarioRunner in these scripts. In EI-Drive, we have developed 16 EIScenarios that cover various challenging driving scenarios, including intersection, traffic lights, overtake, car following, and more. The ego vehicle with simple lane follow policy fails to handle the potential risks in these EIScenarios, leading to crashes. These EIScenarios are valuable for testing and evaluating algorithms in complex driving conditions. Users can also define their own EIScenarios or extend the them by existing OpenSCENARIO scenarios.

\subsection{Modular pipeline}\label{sub:pipeline}
Modular pipeline integrates the sensing, perception, planning, and control modules that enable the vehicle to receive information from its surrounding environment and make decisions. This pipeline establishes a completed data flow from the environment to control, serving as the backbone of the simulation that drives all agents. The four modules in the pipeline are highly decoupled, which enhances their extensibility for incorporating customized methods and enables users to delve deeper into the interactions between pipeline modules and other modules, such as communication model. The following section will detail each module and the interactions between them.

\textbf{Sensing.} Sensing module collects raw data from environment with various sensors, including RGB cameras, LiDAR, GPS, and IMU. These multi-modal sensors gather comprehensive information about environment, constituting the foundation of decision-making for downstream modules.

EI-Drive integrates multiple sensors on an agent, allowing users to customize the type, number, transformation, and other specific parameters of the sensors, such as camera's Field of View (FOV) and LiDAR intensity. This flexibility enables user to design agents with varying sensing capabilities for their experiment. Also, to simplify the sensing process or avoid errors, EI-Drive also provides the oracle method to fetch the data directly from the CARLA server, ensuring data accuracy. For instance, the localization function determines the location and speed of the ego vehicle via a Kalman Filter based on GPS and IMU data. Alternatively, the oracle method can provides accurate location and speed from the server, enabling user to exclude the possible error in sensing and conduct experiments more conveniently and precisely.

\textit{Interaction.} The sensing module closely interacts with simulation environment, where the occluder in the environment influence sensor performance. For example, the heavy traffic flow and large obstacles can obstruct LiDAR and RGB camera, and the adverse weather conditions like heavy rain, fog, and limited sunlight, significantly impair the performance of RGB cameras. As the first section of modular pipeline, the sensing module transmits thees environmental influences throughout the entire pipeline, impacting the vehicle's decisions and actions.

\textbf{Perception.} Perception module processes the raw data from sensing module and interprets it into an understandable representation for downstream modules. In EI-Drive, perception module involves object detection, cooperative perception, traffic sign detection, and traffic light detection. 

The object detection function identifies vehicles and pedestrians from raw data and generates 3D bounding boxes, based on which the subsequent module plans a collision-free trajectory for the vehicle. In EI-Drive, we integrate several classic 2D visual object detection models, including YOLOv5 and Single-Shot Detector (SSD) \cite{liu2016ssd}. These models detect the vehicle and traffic signs in the image captured by RGB cameras and produce 2D bounding boxes. To turn them into 3D bounding boxes, LiDAR point cloud data is integrated to generate 3D detected object in simulation world, which is the output of the perception module. To conduct accurate experiment in perception, we also design an oracle method in perception module, where the 2D and 3D objects within a certain range of the ego vehicle are obtained and visualized based on oracle data.

Cooperative perception function aims to extend ego vehicle's perception capability by combining the perception results from other agents, including vehicles and RSUs. EI-Drive enables the fusion of perception data from multiple resources and visualizes the results on another interface using Pygame. The built-in fusion method, as a part of edge-AI module, combines the bounding boxes of the same object, serving as a simple example method in cooperative perception. Particularly, the perception module visualizes the cooperative perception results, while the data fusion is handled by the edge-AI module, which will be discussed in the later section.

\textbf{Planning.} The planning module is responsible for determining a collision-free and efficient trajectory for the agent. The planning involves global planning, behavior planning, and trajectory planning sequentially.

The global planning is based on A* \cite{hart1968formal}, a classic path-finding algorithm, which generates a high-level and most efficient path on CARLA road map. Global planning does not account for the detailed movements of agents, while the behavior planning determines the specific movements required to follow on the global path. In behavior planning, we define the vehicle behavior to handle situation such as overtake, car-following, lane change and traffic signal. Based on the perception results from upstream modules, behavior planning helps the ego vehicle decide whether it is feasible to take a specific action. For example, when there are obstacles on the adjacent lane during an overtaking maneuver, behavior planning stops overtaking and switches to car-following to ensure safety. Finally, trajectory planning synthesizes the results of global planning and behavior planning, producing a collision-free and smooth trajectory.

\textit{Interaction.} The trajectory adapts to dynamic environment, enabling vehicle to avoid collision and follow traffic signal timely. When obstacles emerge ahead, vehicle reacts appropriately by overtaking, following, or braking, depending on the relative speed. Real-time planning tightly connects vehicle with dynamic environment that any changes in simulation world may lead to different decisions made by the vehicle. Additionally, since the planning is based on the perception results, its performance is heavily reliant on the performance of object detection. Undetected obstacles, whether due to limited visibility or instability in perception, can result in planning failures and potential crashes. This highlights that the performance of autonomous driving algorithm, as a pipeline, significantly depends on the performance of each component.

\textbf{Control.} The control module maneuvers vehicle by throttle, braking, and steering to follow the trajectory generated by the planning module. The kernel of the control module is a controller, assisted by rule-based instructions and a converter.

The controller generates actions, such as acceleration and steering angle, based on the trajectory and desired speed. EI-Drive employs a Proportional-Integral-Derivative (PID) controller, a simple yet stable controller for trajectory following. Beyond controller, user can apply rule-base instructions that take precedence over the controller. For example, a stop mode is available for each vehicle, which, when activated, overrides the outputs of controller and halts the vehicle. Also, in the experiment where the vehicle loses partial steering capacity under specific conditions, rule-based instructions can be defined in the control module to fulfill the requirement. This approach offers more flexibility for testing and debugging.

\subsection{Edge-AI module}\label{sub:edge}
Edge-AI module is designed to simulate the communication and data processing among edge devices, including vehicles and RSUs. The module primarily consists of two key functions: communication model and data fusion.

\textbf{Communication model.} The communication model simulates the characteristics of the communication between agents and its influence on the pipeline. The model addresses two critical quality metrics in communication system: latency and error. The transmission latency represents a time delay between when the data is sent by one agent and when it is received by the ego vehicle. Long latency causes ego vehicle to make decisions based on outdated data, potentially leading to crashes. The variation of the latency can be set as either deterministic or stochastic in the simulation. The transmission error indicates the frame loss due to unstable channel conditions or system errors. The missing frame results in the ego vehicle losing part of information, which impairs the performance. The error rate can be set to adjust the probability of frame loss.

The communication model is applied to perception result received from other agents to the ego vehicle in cooperative perception. Due to transmission latency, the bounding boxes received represent the objects as they were a short time before. Additionally, the ego vehicle may lose some frames due to transmission errors. When the ego vehicle's perception results are not timely or complete, the overall performance of autonomous driving is degraded.

\begin{table*}[t]
   \caption{The settings of experiment scenarios.}  
   \label{tab:intro}
   \normalsize
   \centering
   \begin{tabular}{cccc}
   \toprule\toprule
   \textbf{Experiments} & \textbf{Scenarios} & \textbf{Description}\\ 
   \midrule
   \multirow{4}{*}{Pipeline Module Test} & Pipeline Scenario 1 & Ego vehicle overtakes two vehicles ahead.\\
                                    & Pipeline Scenario 2 & Ego vehicle follows the vehicle ahead and keep a safe distance.\\
                                    & Pipeline Scenario 3 & Ego vehicle deals with the traffic light.\\
                                    & Pipeline Scenario 4 & Ego vehicle deals with the stop sign.\\
   \midrule
   \multirow{4}{*}{\makecell{Cooperative Perception \\ in Collision Avoidance}} & Coop. Scenario 1 & Collision avoidance with a RSU.\\
                                    & Coop. Scenario 2 & Collision avoidance with a spectator vehicle.\\
                                    & Coop. Scenario 5\&6 & Collision avoidance with a RSU and a spectator vehicle.\\
                                    & Coop. Scenario 7\&8 & Collision avoidance with a RSU and a spectator vehicle.\\
   \midrule
   \multirow{2}{*}{\makecell{Cooperative Perception \\ in Object Detection}} & Coop. Scenario 3 & Object detection on the traffic flow with spectator vehicles.\\
                                    & Coop. Scenario 4 & Object detection on the traffic flow with RSUs.\\
   \bottomrule
   \end{tabular}
\end{table*}

\textbf{Data fusion.} Data fusion module is responsible for integrating and processing data from multiple sources. A practical application is the data fusion in cooperative perception, where the module receives bounding boxes from multiple spectators and combines overlapping bounding boxes of the same object to generate stable bounding boxes for each object. In this case, we utilized a simple late fusion method to distinguish and average the bounding boxes, while the data fusion method can be customized by any other algorithms for further research.

An key data source in this process is the RSU, which can be installed at elevated position with better visibility in critical traffic scenarios. Additionally, RSUs typically have superior sensors, computing power, and communication components, allowing them to provide high-quality data to all agents nearby and significantly enhance the efficiency and safety. A essential function of the data fusion module is utilizing the data from RSUs to improve ego vehicle's driving performance. We have also conducted extensive experiments to illustrate the improvement achieved through data fusion and the participation of RSUs, which will be discussed in the experiment section.

\textit{Interaction.} The communication model direct impacts the perception module by applying transmission latency and errors on the perception results received by vehicles. Since the subsequent decisions heavily depends on the perception results, the latency and errors can have significant negative effect on driving performance. Furthermore, the communication model can be extended to standalone perception experiments, where all perception results originate from the ego vehicle itself. In this context, latency and errors are attributed to data processing latency and sensor errors, respectively. Although the communication model is not originally designed to simulate data processing latency and sensor errors, they have equivalent effects in some conditions, allowing the platform to support a broader range of research.

\section{Experiments}
In this section, extensive experiments are conducted to showcase the key features of EI-Drive and compare the performance across different settings. In \Cref{subsec:pipeline experiment}, we illustrate the features of pipeline modules by four different scenarios. In \Cref{subsec: coop collision experiment}, we highlight the role of cooperative perception in collision avoidance tasks and demonstrate that it enhances vehicle safety. In \Cref{subsec: coop object experiment}, we explore cooperative perception in object detection tasks and compare its performance in various settings.

\textbf{Experiment settings.} To better illustrate the features of EI-Drive and enhance the validity of the experiment, we design several scenarios for each experiment below with various settings. The experiment scenarios are detailed in the \Cref{tab:intro}. In all scenarios, the spectator vehicles and the RSUs are equipped with the same RGB cameras and LiDAR, with RSUs installed at appropriate heights.

\subsection{Pipeline module} \label{subsec:pipeline experiment}
We design four different tasks, including overtaking, vehicle following, traffic light, and stop sign, to demonstrate the pipeline module's capability to make appropriate driving decisions and handle various driving scenarios. In all the scenarios, the ego vehicle utilizes the multi-modal sensors as input and employs various perception methods.

\begin{figure}[!t]
\centering
\subfloat[Pipeline scenario 1: overtake.\label{fig:1a}]{\includegraphics[width=1.0\linewidth]{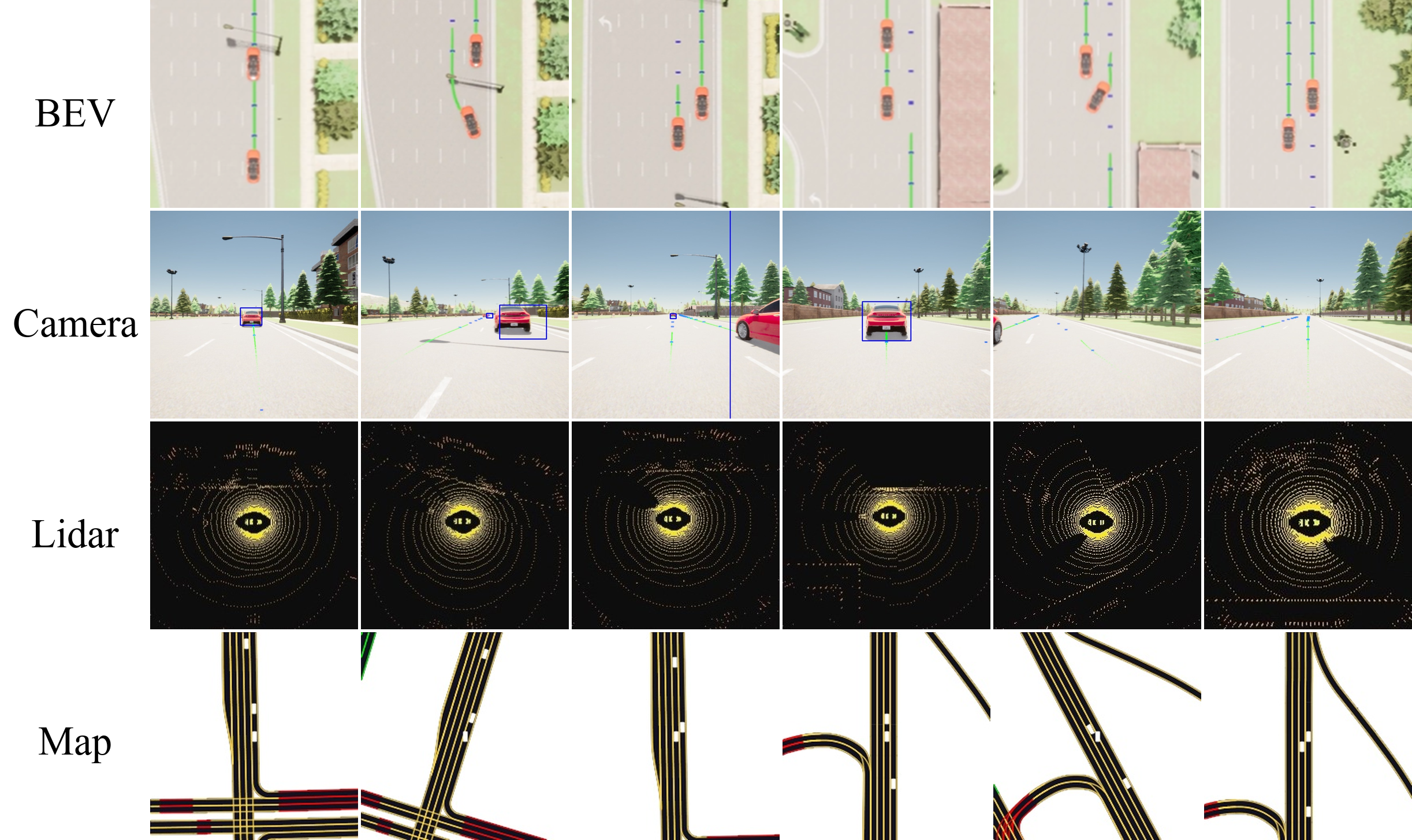}%
\label{fig:1}}
\vspace{-0.8em}
\vfil
\subfloat[Pipeline scenario 2: vehicle following.\label{fig:1a}]{\includegraphics[width=1.0\linewidth]{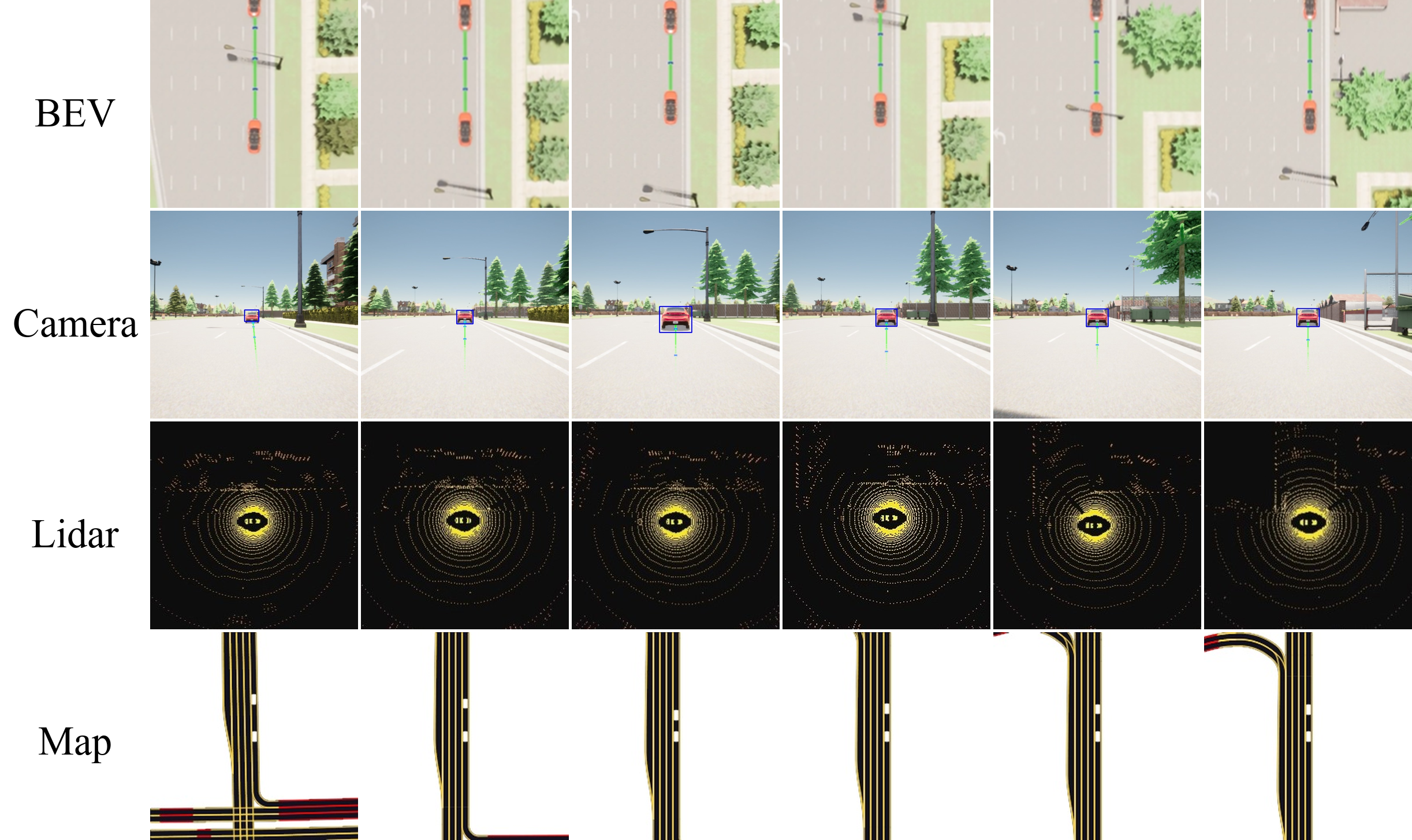}%
\label{fig:1}}
\vfil
\caption{Multi-modal sensors in pipeline scenarios.}
\label{fig:multi-modal}
\end{figure}

\textbf{Multi-modal sensors.} \Cref{fig:multi-modal} illustrates ego's behavior in each scenario using multi-modal sensors. In the overtake scenario, ego vehicle successfully overtakes two vehicles ahead consecutively, ensuring lane changes are made safely without collision. In the vehicle following scenario, the ego vehicle follows the vehicle in front with a safe distance. In the traffic light scenario, the ego vehicle properly navigates the traffic light and passes through the intersection. In the stop sign scenario, the ego vehicle stops in front of the stop sign and then proceeds forward. The experiment demonstrates that the ego vehicle can detect objects in the environment and response correctly using the pipeline of EI-Drive, ensuring safe and smooth trajectory planning and movement control.

\begin{figure}[!t]
\centering
\subfloat[Pipeline scenario 1: overtake.\label{fig:1a}]{\includegraphics[width=1.0\linewidth]{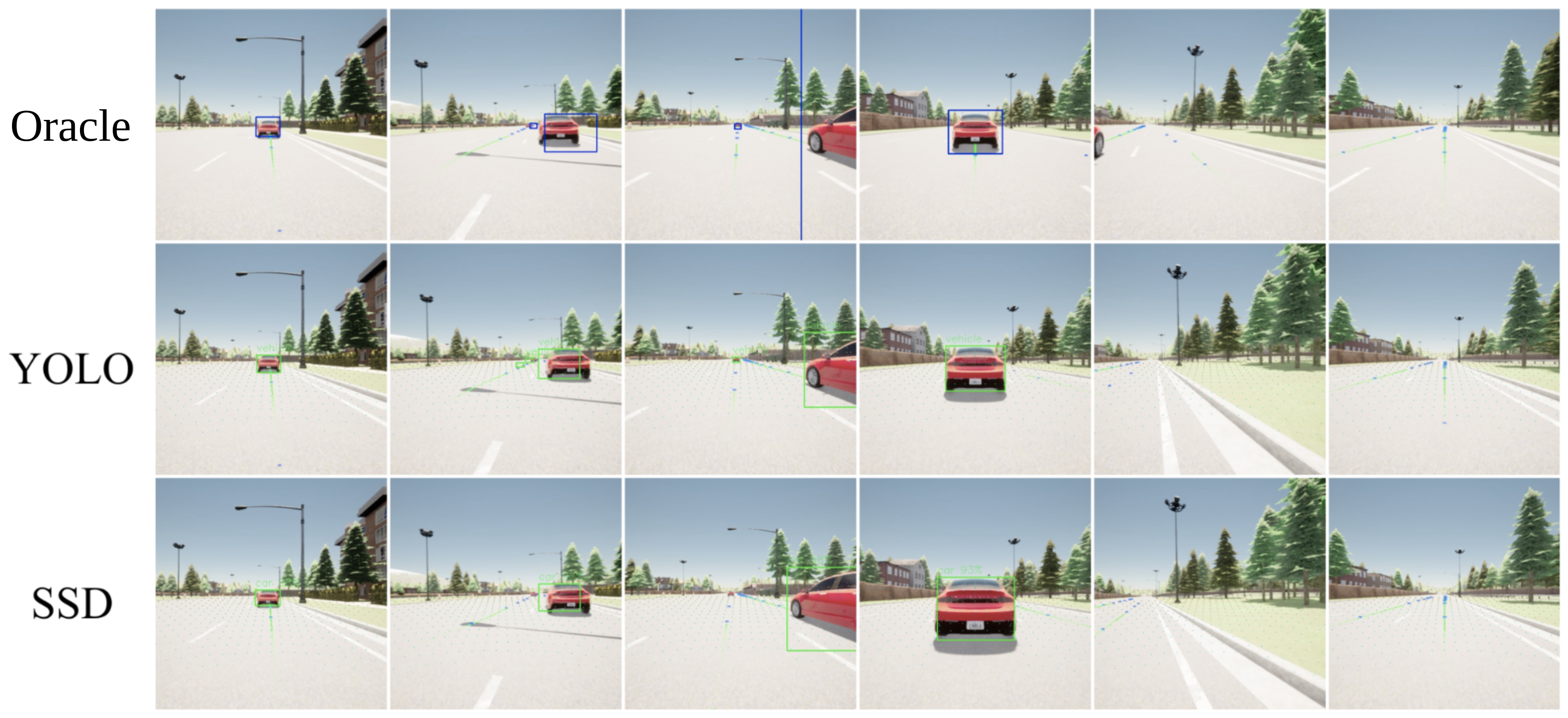}%
\label{fig:1}}
\vspace{-0.8em}
\vfil
\subfloat[Pipeline scenario 2: vehicle following.\label{fig:1a}]{\includegraphics[width=1.0\linewidth]{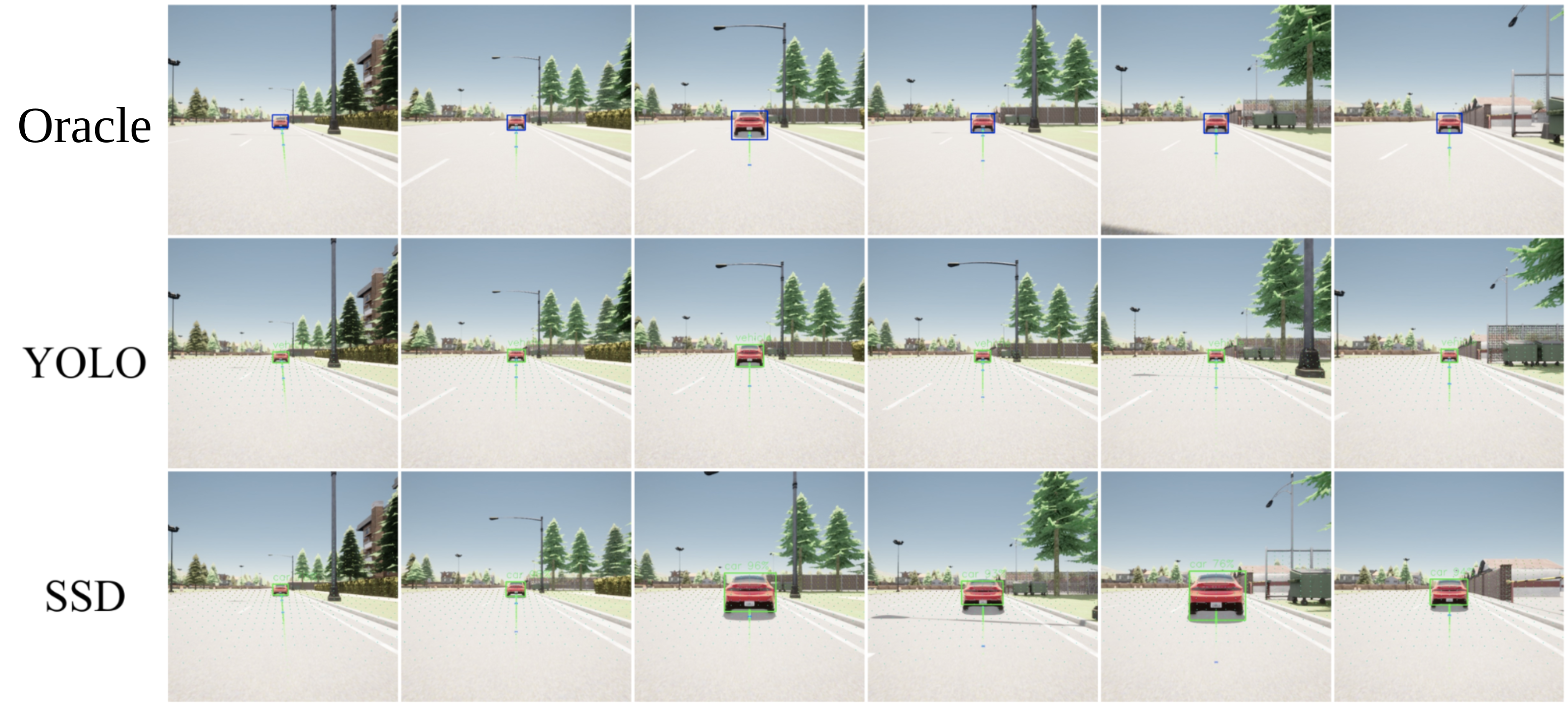}%
\label{fig:1}}
\vfil
\caption{Various object detection methods in pipeline scenarios.}
\label{fig:perception method}
\end{figure}

\textbf{Perception methods.} To further detail the object detection methods in the perception module, we test the built-in perception methods in the overtaking and vehicle following scenarios. The experiment utilizes the built-in oracle, YOLOv5, and SSD for object detection on the camera images of the ego vehicle. \Cref{fig:perception method} shows that these methods accurately detect the vehicles and visualize the bounding boxes.

\subsection{Cooperative perception in collision avoidance} \label{subsec: coop collision experiment}
To better highlight that cooperative perception extends the ego vehicle's perception capability and enhance safety, we design the experiments on collision avoidance tasks. In the Coop. Scenario 1 \& 2, the ego vehicle navigates an intersection without traffic lights, while another incoming vehicle from the left, obscured by a firetruck, poses a potential collision. To avoid the collision, cooperative perception brings extra information about this hidden vehicle from either a spectator vehicle or RSU to help the ego vehicle. Since the experiment focuses on perception module, we simplify the planning module by stopping the ego vehicle whenever the incoming vehicle is detected within a specific area near intersection, reducing the perturbations from other modules. Therefore, the key to the collision avoidance is whether the ego vehicle can detect the approaching vehicle in advance.

\begin{figure}[!t]
\centering
\subfloat[Coop. scenario 1.\label{fig:1a}]{\includegraphics[width=1.0\linewidth]{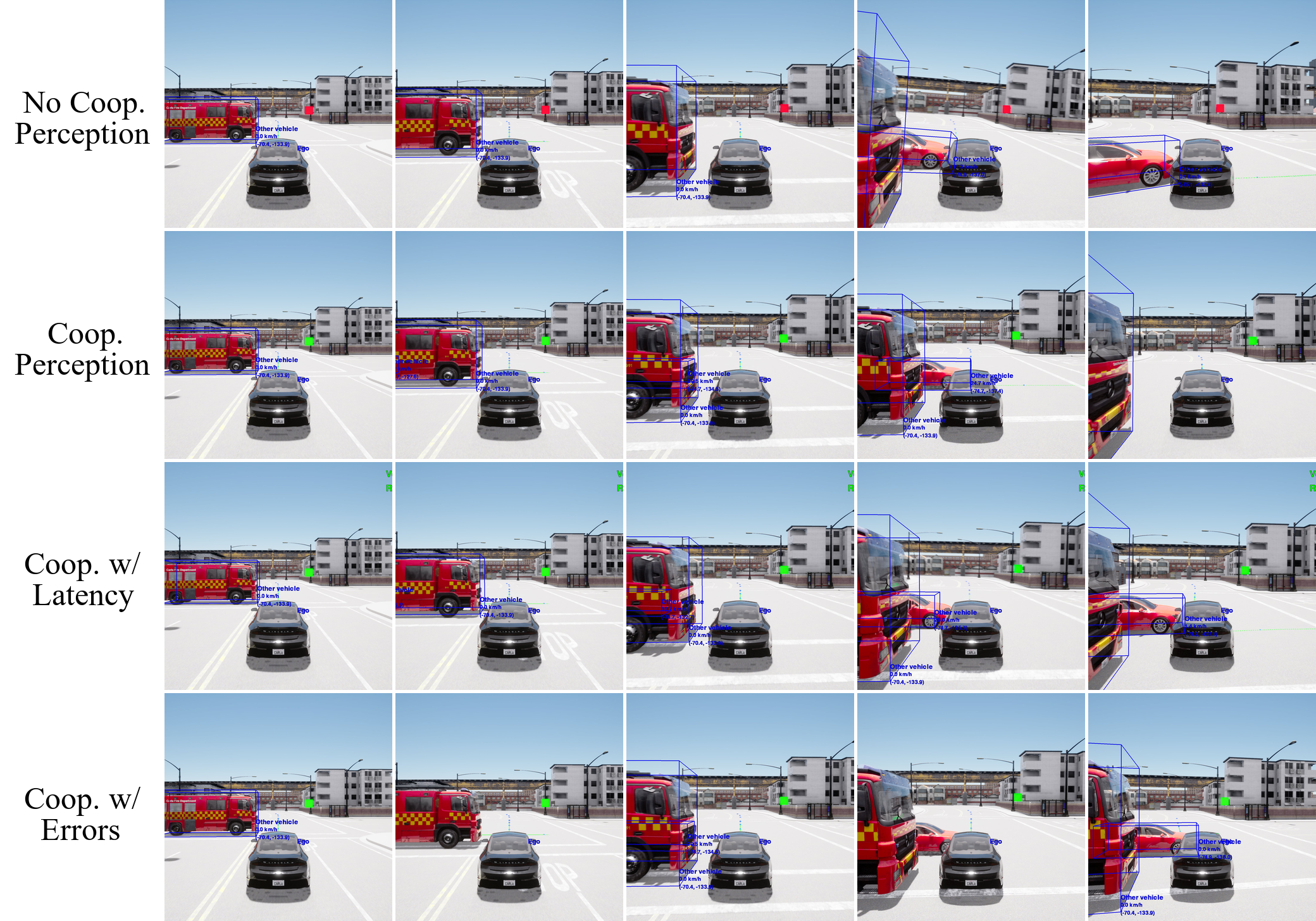}}
\vspace{-0.8em}
\vfil
\subfloat[Coop. scenario 2.\label{fig:1a}]{\includegraphics[width=1.0\linewidth]{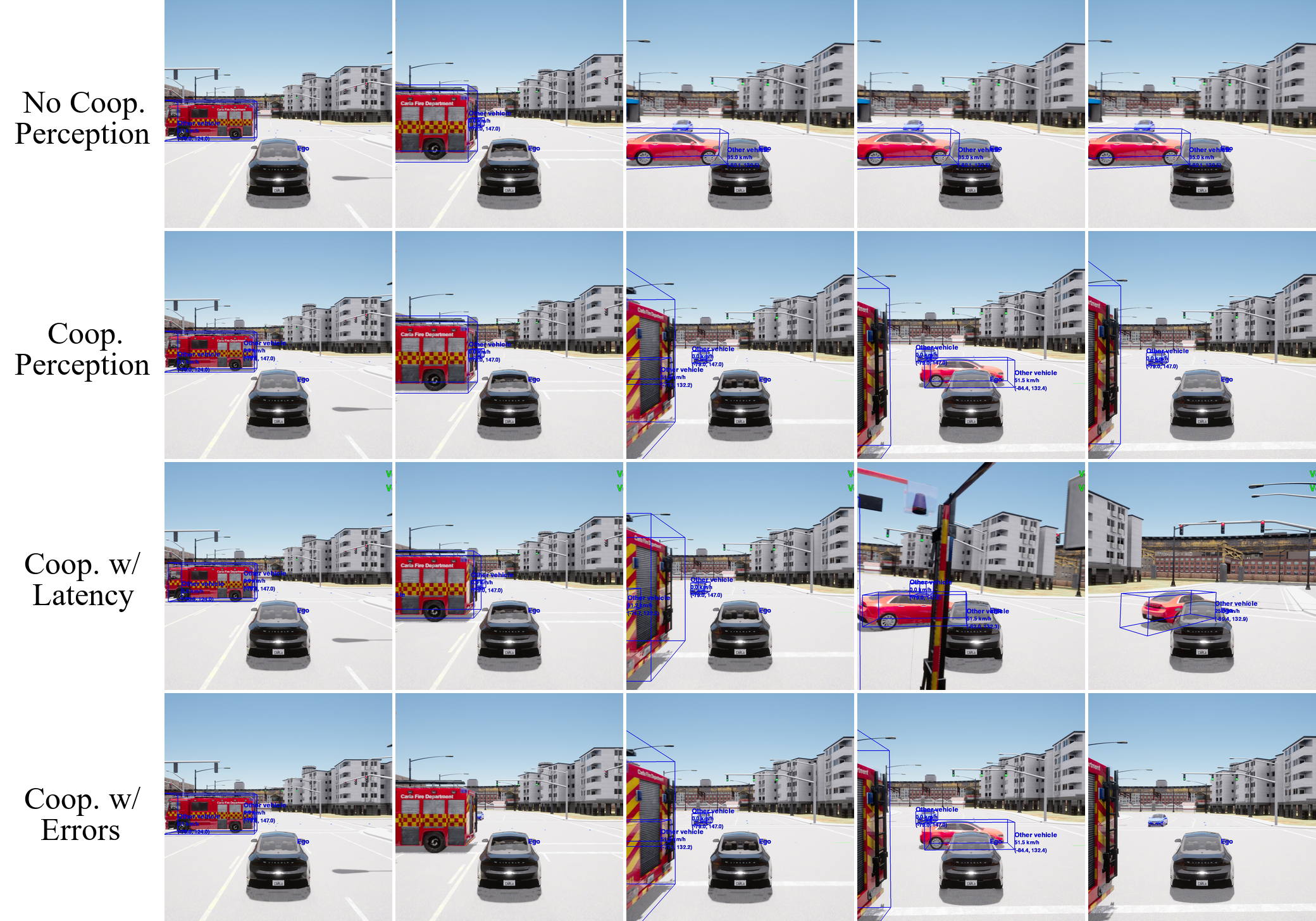}}
\vfil
\caption{Cooperative perception in collision avoidance tasks by oracle method.}
\label{fig:coop in ca}
\end{figure}

\Cref{fig:coop in ca} presents the detail of the experiments, where we study the influence of cooperative perception, transmission latency, and transmission errors. In this experiment, we set the latency and error rate to 0.3 seconds and 30\%, respectively. As shown in the figure, the bounding box of the incoming vehicle behind the firetruck is visualized only when cooperative perception is enabled. However, without cooperative perception, the incoming vehicle is only detected when it comes into view of the ego vehicle from behind the firetruck, while it is too late for the ego vehicle to brake. The third and fourth rows highlight the effects of transmission latency and error, which results in deviation of the bounding box positions from their ground truth position and the loss of some the bounding box frames, respectively.

To precisely quantify these effects, we repeat the experiment a sufficient number of times and calculate relevant metrics, which are recorded in \Cref{tab:coop 1&2}. The experiments are undertaken with two perception methods: oracle and YOLOv5. The minimal distance in the table represents the minimal distance between the ego vehicle and the incoming vehicle in a complete episode. A smaller minimal distance indicates poorer safety for the ego vehicle.

The results in the first and second rows of the table show that the ego vehicle keeps a safer distance from the incoming vehicle with cooperative perception enabled. It demonstrates that cooperative perception allows the ego vehicle to detect the obstructed incoming vehicle and brake in advance, while it cannot be achieved by the ego vehicle alone. Comparing the second row with the third and fourth rows validates that the transmission latency and errors significantly impair the performance of cooperative perception.

Furthermore, to highlight the benefits from combining the data from both the spectator vehicle and the RSU, we design Coop. Scenario 5-8 and carry out the collision avoidance experiments to provide comprehensive results. The key difference between these four scenarios and the Coop. Scenario 1 \& 2 is that both the RSU and the spectator vehicle are present in Coop. Scenario 5-8, allowing the ego vehicle to receive information from both sources. Heavy traffic flow is also introduced in Cooperative Scenarios 5–8 to enhance the realism of the experiment. In this experiment, we vary the source of cooperative perception data to compare the perception capabilities between the vehicle and the RSU. The success rate of experiment is defined in \Cref{eq:success_rate} and recorded as an important metric. Here, $N_{cf}$ and $N_{total}$ represents the number of collision-free attempts and total attempts, respectively.

\begin{equation}
\text{Success Rate} = \frac{N_{cf}}{N_{total}} \times 100\%
\label{eq:success_rate}
\end{equation}

As shown in \Cref{tab:coop 5-8}, the success rate is significantly higher when both RSU and the spectator vehicle participate in cooperative perception, while it is notably lower when only the spectator vehicle is involved. This difference is due to the heavy traffic flow obstructing the view of the spectator vehicle. This demonstrates that the view of the vehicle is limited by heavy traffic, whereas the RSU has a significant advantage in perception since it consistently offers a wide and unobstructed view from its elevated position. 

\subsection{Cooperative perception in object detection} \label{subsec: coop object experiment}
To further validate the performance of cooperative perception across different tasks, we design the Coop. Scenario 3 and 4 and undertake the experiment. In these scenarios, the ego vehicle aims to detect the vehicles in the heavy traffic with the help of spectator vehicles and RSUs at the intersection, respectively. The number of vehicle detected serves as an important metric for evaluating the performance, which is constrained by both view and detection range.

\begin{figure}[!t]
\centering
\subfloat[Coop. scenario 3.\label{fig:coop in od 3}]{\includegraphics[width=1.0\linewidth]{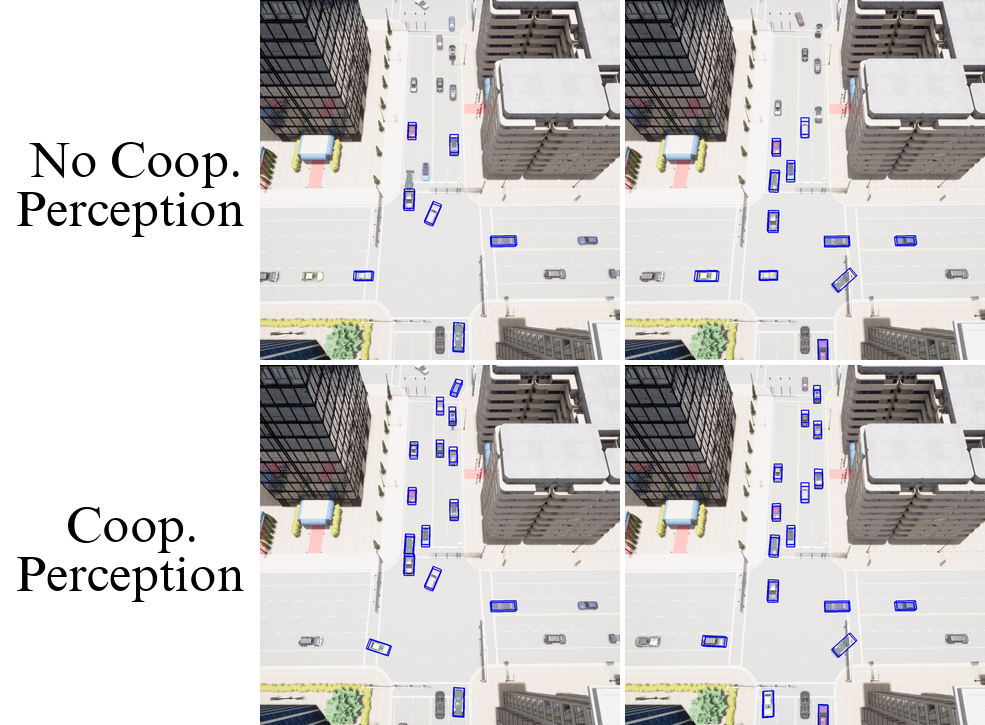}}
\vspace{-0.0em}
\vfil
\subfloat[Coop. scenario 4.\label{fig:coop in od 4}]{\includegraphics[width=1.0\linewidth]{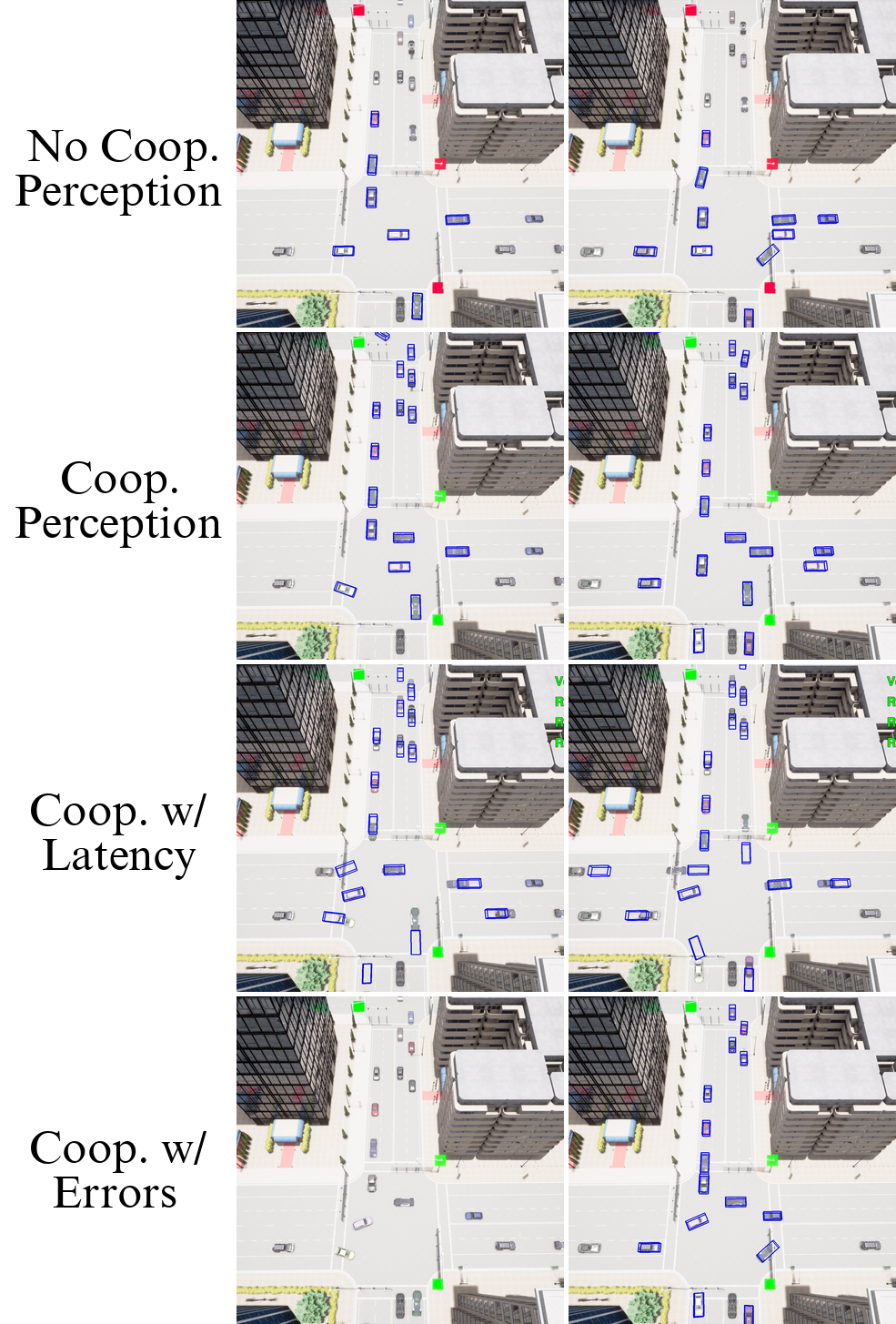}}
\vfil
\caption{Cooperative perception in object detection tasks by oracle method.}
\label{fig:coop in od}
\end{figure}

As shown in the perception result in \Cref{fig:coop in od}, the ego vehicle's perception range is limited without cooperative perception since its view is obstructed by surrounding vehicles. When cooperative perception is enabled, the spectator vehicles in \Cref{fig:coop in od 3} and the RSUs in \Cref{fig:coop in od 4} share the perception information to extend the perception range, greatly increasing the number of vehicles detected. Additionally, given that the object detection is not always stable, redundant detection from multiple sources on the same objects improves the robustness of the perception. Consequently, cooperative perception greatly enhances the perception capability in urban heavy traffic.

\begin{figure}[!t]
\centering
\subfloat[Coop. scenario 3.\label{fig:od1}]{\includegraphics[width=1.0\linewidth]{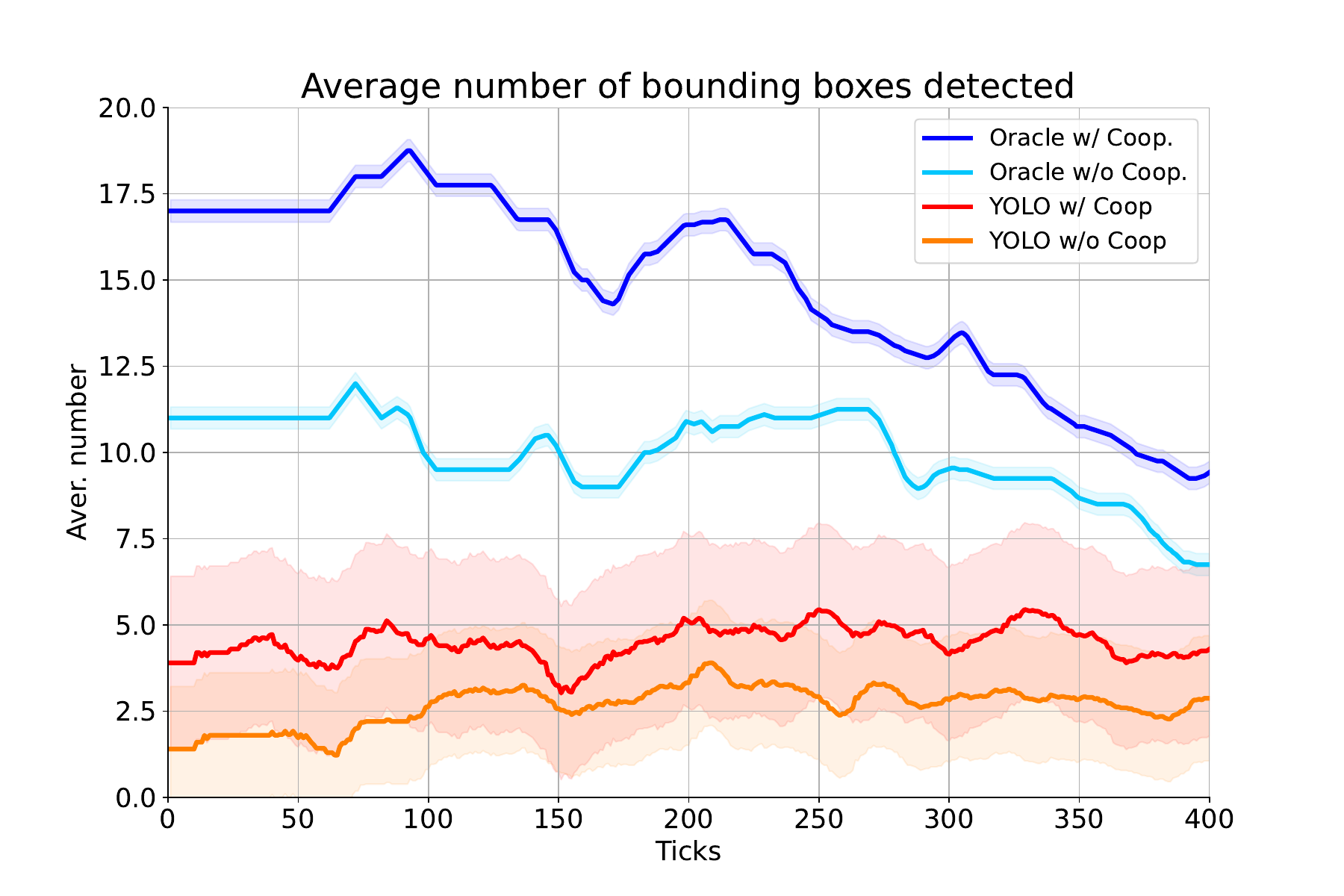}}
\vspace{-0.5em}
\vfil
\subfloat[Coop. scenario 4.\label{fig:od2}]{\includegraphics[width=1.0\linewidth]{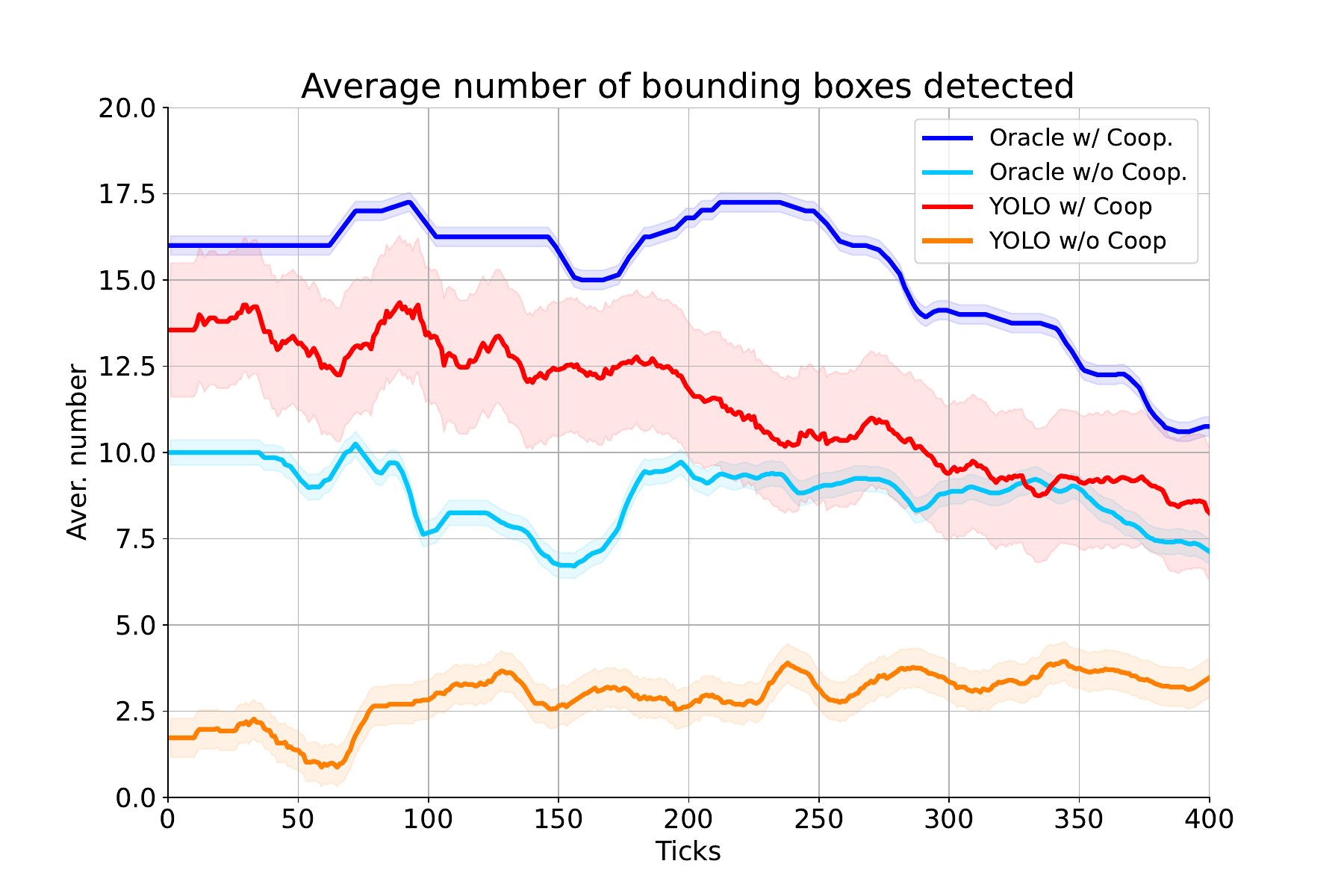}}
\vspace{-0.5em}
\vfil
\subfloat[Comparison between the spectator vehicle and RSU.\label{fig:od3}]{\includegraphics[width=1.0\linewidth]{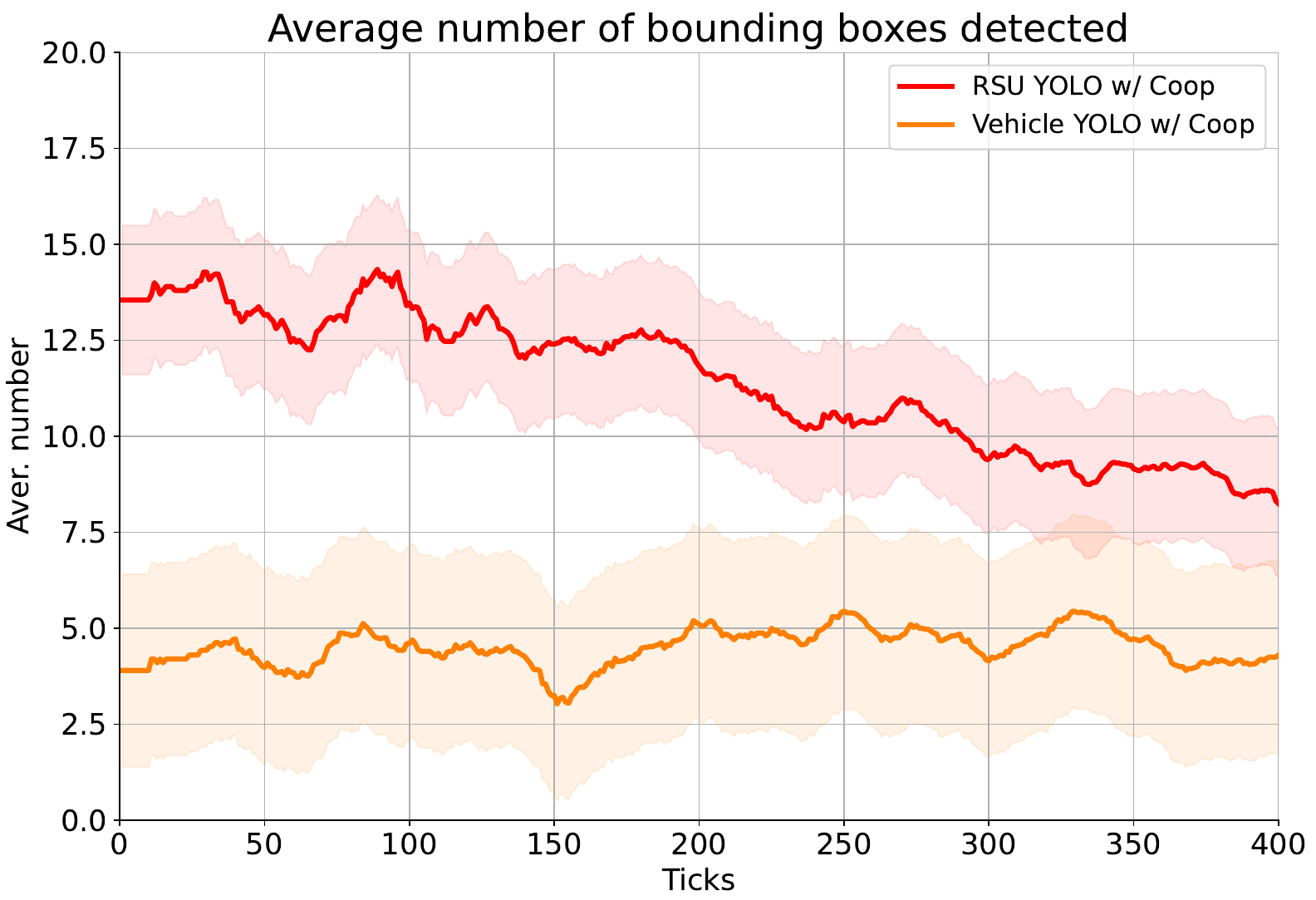}}
\vfil
\caption{The number of detected object under different settings.}
\label{fig:number of object detection}
\end{figure}

We adopt the number of detected objects as a metric and quantify the results in \Cref{fig:number of object detection}. In \Cref{fig:od1} and \Cref{fig:od2} the cooperative perception shows a significant advantage over standalone perception in both scenarios with different perception methods. To compare the perception capabilities of the spectator vehicle and the RSU, we present \Cref{fig:od3} with the YOLOv5 detection results from Coop. Scenario 3 and 4. Under the same traffic condition, the RSUs detect more vehicles due to its unobstructed view.

\begin{table*}[t]
   \caption{Minimal distance between the ego vehicle and the incoming vehicle in Coop. scenario 1 \& 2.}  
   \label{tab:coop 1&2}
   \large
   \centering
   \begin{tabular}{cccccc}
   \toprule\toprule
   & \multicolumn{2}{c}{\textbf{Coop. Scenario 1}} & \multicolumn{2}{c}{\textbf{Coop. Scenario 2}} \\ 
   \cmidrule(lr){2-3} \cmidrule(lr){4-5}
    & \textbf{Oracle} & \textbf{YOLOv5} & \textbf{Oracle} & \textbf{YOLOv5} \\ 
   \midrule
   No Coop. perception & 3.26 $\pm$ 0.08 & 3.32 $\pm$ 0.00 & 3.90 $\pm$ 0.01 & 3.91 $\pm$ 0.00 \\
   Coop. perception & 3.98 $\pm$ 0.16 & 5.09 $\pm$ 0.13 & 4.42 $\pm$ 0.00 & 4.58 $\pm$ 0.04\\
   Coop. perception w/ latency & 2.50 $\pm$ 0.68 & 3.62 $\pm$ 0.08 & 3.50 $\pm$ 0.01 & 3.40 $\pm$ 0.00\\
   Coop. perception w/ error & 3.15 $\pm$ 1.07 & 4.84 $\pm$ 0.55 & 4.32 $\pm$ 0.21 & 4.23 $\pm$ 0.49\\
   \bottomrule
   \end{tabular}
\end{table*}

\begin{table*}[t]
   \caption{Metrics in Coop. scenario 5 - 8 with different data fusion settings.}  
   \label{tab:coop 5-8}
   \large
   \centering
   \begin{tabular}{cccc}
   \toprule\toprule
   \textbf{Scenarios} & \textbf{Participants} & \textbf{Success Rate} & \textbf{Min. Distance(m)}\\ 
   \midrule
   \multirow{3}{*}{Coop. Scenario 5} & Vehicle + RSU & 100.00\% & 3.51 $\pm$ 0.55 \\
                                    & RSU & 93.33\% $\pm$ 4.71\% & 3.03 $\pm$ 0.03\\
                                    & Vehicle & 20.51\% $\pm$ 3.63\% & 2.98 $\pm$ 0.48\\
   \midrule
   \multirow{3}{*}{Coop. Scenario 6} & Vehicle + RSU & 82.05\% $\pm$ 7.25\% & 4.05 $\pm$ 1.07 \\
                                  & RSU & 78.46\% $\pm$ 1.66\% & 3.10 $\pm$ 0.08\\
                                  & Vehicle & 48.72\% $\pm$ 3.63\% & 3.32 $\pm$ 0.63\\
    \midrule
   \multirow{3}{*}{Coop. Scenario 7} & Vehicle + RSU & 100\% & 3.83 $\pm$ 0.11 \\
                                  & RSU & 98.33\% $\pm$ 2.36\% & 3.60 $\pm$ 0.00\\
                                  & Vehicle & 0.00\% & 2.82 $\pm$ 0.03\\
    \midrule
   \multirow{3}{*}{Coop. Scenario 8} & Vehicle + RSU & 100.00\% & 3.60 $\pm$ 0.00 \\
                                  & RSU & 100.00\% & 3.61 $\pm$ 0.01\\
                                  & Vehicle & 0.00\% & 2.88 $\pm$ 0.04\\
   \bottomrule
   \end{tabular}
\end{table*}

\section{Conclusion}
\label{sec:conclusion}

In this paper, we present EI-Drive, an platform for cooperative perception with realistic communication models. Particularly, we integrate realistic communication models that account for transmission latency and errors with autonomous driving pipeline, enabling the research on their impact on cooperative perception. The platform includes built-in customizable driving scenarios tailored for evaluating algorithm under complex traffic and realistic network conditions, bridging the gap in autonomous driving simulators and advancing autonomous driving development. The experiments performed on EI-Drive cover various cooperative perception tasks under diverse network conditions, highlighting that the transmission latency and errors impair the overall performance of cooperative perception.

Looking ahead, leveraging EI-Drive presents a promising approach to developing autonomous driving algorithms that are robust to realistic network conditions. Given that the data transmission is a crucial aspect of real-world vehicular networks, training algorithms under complex network conditions can significantly improve their robustness against the negative effects of latency and errors, which fosters the real-world implementation of autonomous driving algorithms.

Furthermore, EI-Drive will benefit from the community contribution. As an open-source platform with high customizability, it encourages contributors to integrate their own components and algorithms, enabling a wide range of research on the platform. This collaborative approach will significantly enrich EI-Drive's utility and advance the development of related research.




\bibliographystyle{IEEEtran}
\bibliography{references}

\newpage

\section{Biography Section}
\begin{IEEEbiography}[{\includegraphics[width=1in,height=1.25in,clip,keepaspectratio]{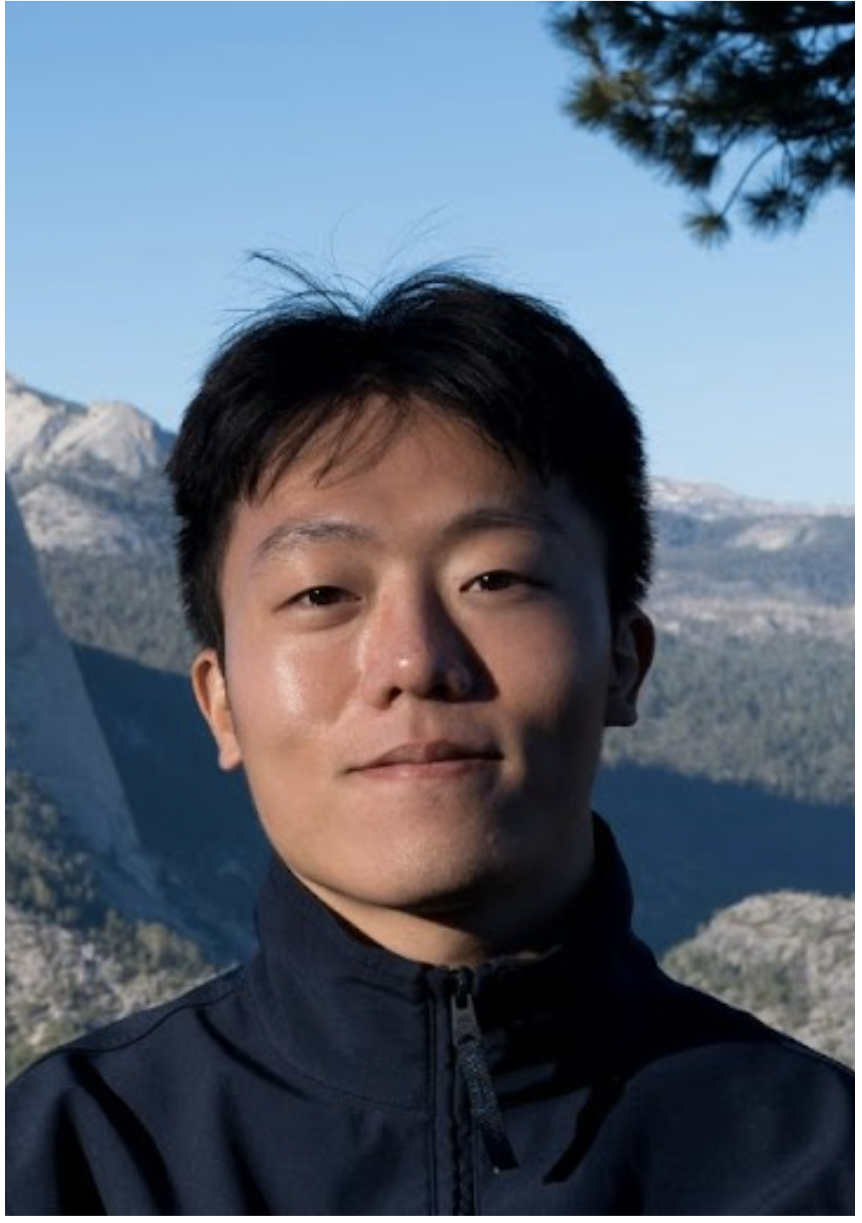}}]{Hanchu Zhou}
Hanchu Zhou received the B.S. degree in Control Science from Zhejiang University, Hangzhou, China. He is currently pursuing the Ph.D. degree in Electrical and Computer Engineering at the University of California, Davis.

His research interests include reinforcement learning and multi-agent systems, aiming at enhancing model generalization capability and efficiency by learning methods. His recent research primarily focuses on world model based autonomous driving.
\end{IEEEbiography}

\begin{IEEEbiography}[{\includegraphics[width=1in,height=1.25in,clip,keepaspectratio]{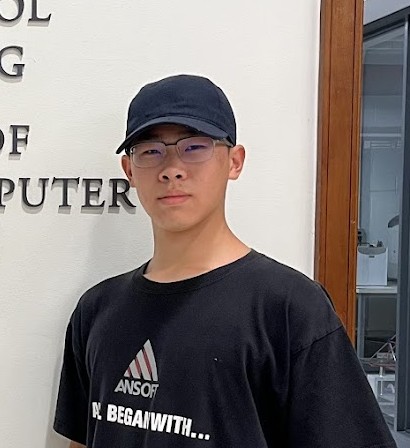}}]{Edward Xie}
Edward Xie is currently an undergraduate in the Whiting School of Engineering at Johns Hopkins University, studying Electrical Engineering and Computer Science. His research interests include practical applications of robotics, sensing, and control and the integration of AI into robotic systems.
\end{IEEEbiography}

\begin{IEEEbiography}[{\includegraphics[width=1in,height=1.25in,clip,keepaspectratio]{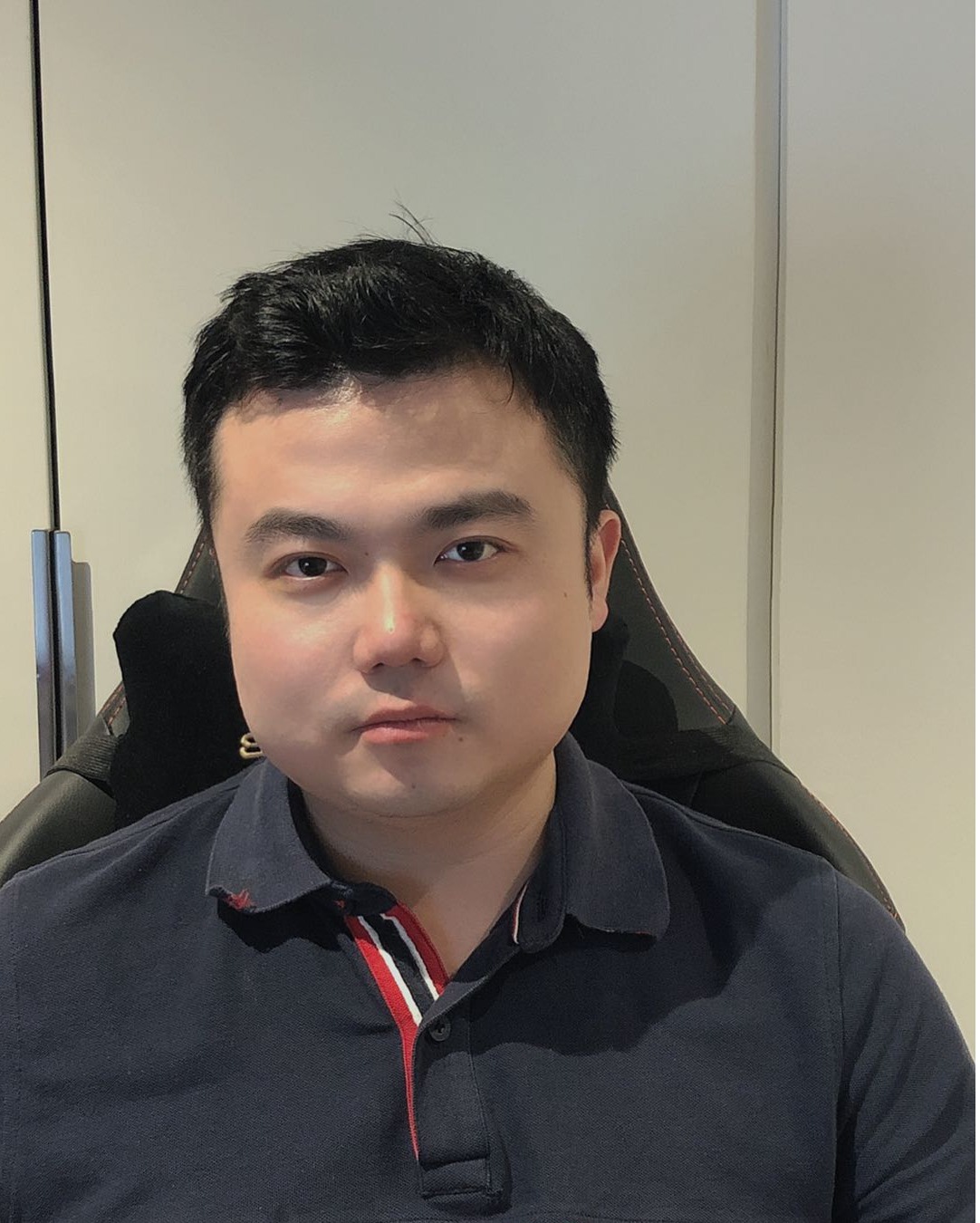}}]{Wei Shao}
Wei Shao is an IWY (Impossible Without You) Research Scientist at CSIRO Data61, an Adjunct Lecturer at the University of New South Wales (UNSW), a Visiting Researcher at the University of California, Davis (UC Davis), and an Adjunct Fellow at RMIT University. Previously, he served as a Postdoctoral Researcher at UC Davis and Arizona State University. His research interests encompass AI-Cybersecurity, Graph Neural Networks, Spatio-temporal Data Mining, Reinforcement Learning Applications, and the Internet of Things (IoT). Wei has authored over 60 research papers published in leading conferences and journals. His contributions to academia have been recognized with accolades such as the Distinguished Paper Award at UbiComp and the Best Reviewer Award at KDD. Additionally, he received CSIRO’s Early Career in Science Award for his outstanding scientific contributions to Australian research.
\end{IEEEbiography}

\begin{IEEEbiography}[{\includegraphics[width=1in,height=1.25in,clip,keepaspectratio]{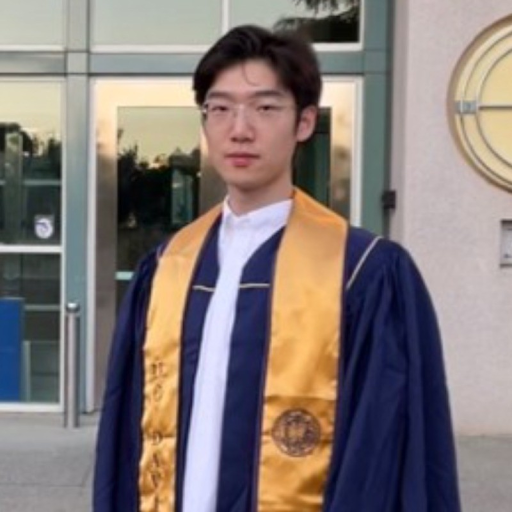}}]{Dechen Gao}
Dechen Gao received the B.S. degree in computer science from Shandong University, Shandong, China, and the M.S. degree in computer science from the University of California, Davis, CA, USA, in 2021 and 2023, respectively, where he is currently pursuing the Ph.D. degree in computer science. He is co-advised by Prof. Iman Soltani and Prof. Junshan Zhang. His research interests include world model-based robotics and autonomous driving. He has interned with Meta Platforms, Inc., Menlo Park, CA, USA, as a Software Development Engineer in 2022, and with TikTok, San Jose, CA, USA, as a Machine Learning Engineer in 2024.
\end{IEEEbiography}

\begin{IEEEbiography}[{\includegraphics[width=1in,height=1.25in,clip,keepaspectratio]{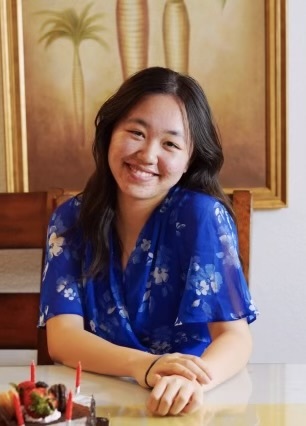}}]{Michelle Dong}
Michelle Dong is currently a senior at Monta Vista High School, USA. Her research interests include reinforcement learning and autonomous driving.
\end{IEEEbiography}

\begin{IEEEbiography}[{\includegraphics[width=1in,height=1.25in,clip,keepaspectratio]{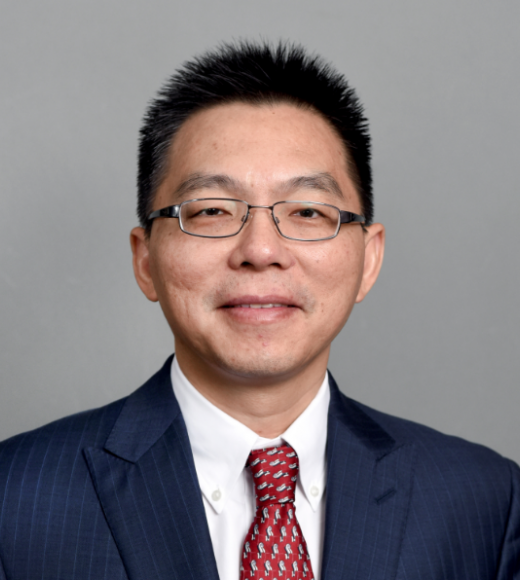}}]{Junshan Zhang}
Junshan Zhang has been a professor in the ECE Department at University of California Davis since 2021. He received his Ph.D. degree from the School of ECE at Purdue University in Aug. 2000, and was on the faculty of the School of ECEE at Arizona State University from 2000 to 2021. His research interests fall in the general field of information networks and data science, including edge AI, reinforcement learning, continual learning, network optimization and control, game theory. He is a Fellow of the IEEE, and a recipient of the ONR Young Investigator Award in 2005 and the NSF CAREER award in 2003.   His papers have won a few awards, including the Best Student paper at WiOPT 2018, the Kenneth C. Sevcik Outstanding Student Paper Award of ACM SIGMETRICS/IFIP Performance 2016, the Best Paper Runner-up Award of IEEE INFOCOM 2009 and IEEE INFOCOM 2014, and the Best Paper Award at IEEE ICC 2008 and ICC 2017.  He  is currently serving as  Editor-in-Chief  for IEEE/ACM Transactions on Networking. He served as Editor-in-Chief for IEEE Transactions on Wireless Communication during 2019-2022.  He was TPC co-chair for IEEE INFOCOM 2012 and ACM MOBIHOC 2015. 
\end{IEEEbiography}

\vspace{11pt}
\vfill

\end{document}